\documentclass[letterpaper, 10 pt, conference]{ieeeconf}  

\IEEEoverridecommandlockouts                              

\usepackage{amsmath} 
\usepackage{amssymb}  
\usepackage[colorinlistoftodos]{todonotes}
\usepackage{hyperref}
\usepackage{units}
\usepackage{multirow}
\usepackage{booktabs}
\usepackage[colorinlistoftodos]{todonotes}
\usepackage{siunitx}
\usepackage{url}

\usepackage[printonlyused]{acronym}

\acrodef{RSL}{Robotic Systems Lab}
\acrodef{COM}{center of mass}
\acrodef{SQP}{sequential quadratic problem}
\acrodef{VMC}{virtual model controller}
\acrodef{w.r.t.}{with respect to}
\acrodef{DOF}{degrees of freedom}
\acrodef{ZMP}{zero moment point}
\acrodef{COP}{center of pressure}
\acrodef{IMU}{inertial measurement unit}
\acrodef{COT}{cost of transport}
\acrodef{JPL}{Jet Propulsion Laboratory}
\acrodef{HAA}{hip adduction/abduction}
\acrodef{HFE}{hip flexion/extension}
\acrodef{KFE}{knee flexion/extension}
\acrodef{ZMP}{zero-moment point}
\acrodef{QP}{quadratic programming}
\acrodef{SQP}{sequential quadratic programming}
\acrodef{WBC}{whole-body controller}
\acrodef{HO}{hierarchical optimization}
\acrodef{NLP}{nonlinear programming}
\acrodef{MPC}{Model Predictive Control}
\acrodef{TO}{Trajectory Optimization}
\acrodef{DARPA}{Defense Advanced Research Projects Agency}
\acrodef{JPL}{Jet Propulsion Laboratory}
\acrodef{RBDL}{Rigid Body Dynamics Library}
\acrodef{RL}{Reinforcement Learning}
\acrodef{LIP}{linear inverted pendulum}
\acrodef{CoM}{Center of Mass}
\acrodef{SLQ}{Sequential Linear Quadratic}
\acrodef{DDP}{Differential Dynamic Programming}
\acrodef{SRBD}{single rigid body dynamics}
\acrodef{EOM}{equations of motion}
\acrodef{SDF}{signed distance field}
\acrodef{ESDF}{euclidean signed distance field}
\acrodef{TSDF}{truncated signed distance field}
\acrodef{CHOMP}{Covariant Hamiltonian Optimization for Motion Planning}
\acrodef{CCE}{compound collision environment}
\acrodef{ROS}{Robot Operating System} 

\usepackage[normalem]{ulem}

\usepackage{accents}

\newcommand{\vth}{\mbox{\boldmath $\theta$}}

\newcommand{\vlambda}{\mbox{\boldmath $\lambda$}}

\newcommand{\vom}{\mbox{\boldmath $\omega$}}

\newcommand{\vf}{\mathbf f}
\newcommand{\vg}{\mathbf g}
\newcommand{\vh}{\mathbf h}

\newcommand{\vn}{\mathbf n}

\newcommand{\vp}{\mathbf p}
\newcommand{\vq}{\mathbf q}
\newcommand{\vr}{\mathbf r}

\newcommand{\vu}{\mathbf u}
\newcommand{\vv}{\mathbf v}

\newcommand{\vx}{\mathbf x}

\newcommand{\vI}{\mathbf I}

\newcommand{\vQ}{\mathbf Q}
\newcommand{\vR}{\mathbf R}

\newcommand{\vT}{\mathbf T}

\newif\ifcorrectingmode
\correctingmodefalse
\usepackage{url}
\usepackage[normalem]{ulem}
\ifcorrectingmode
	\newcommand{\revision}[1]{\textcolor{blue}{#1}}
	\newcommand{\deleted}[1]{\textcolor{red}{\ifmmode\text{\sout{\ensuremath{#1}}}\else\sout{#1}\fi}}
	\newcommand{\deletedequation}[2]{\textcolor{red}{\centerline{Removed equation (#1)}}}
\else
	\newcommand{\revision}[1]{#1}
	\newcommand{\deleted}[1]{}
	\newcommand{\deletedequation}[2]{}
\fi

\title{
Collision-Free MPC for Legged Robots in Static and Dynamic Scenes
}

\author{Magnus Gaertner, Marko Bjelonic, Farbod Farshidian, Marco Hutter%
\thanks{All authors are with the Robotic Systems Lab, ETH Z\"urich, 8092 Z\"urich, Switzerland.
{\tt\footnotesize firstname.lastname@mavt.ethz.ch}}%
}

\begin{document}

\maketitle
\begin{abstract}
We present a model predictive controller (MPC) that automatically discovers collision-free locomotion while simultaneously taking into account the system dynamics, friction constraints, and kinematic limitations. A relaxed barrier function is added to the optimization's cost function, leading to collision avoidance behavior without increasing the problem's computational complexity. Our holistic approach does not require any heuristics and enables legged robots to find whole-body motions in the presence of static and dynamic obstacles. We use a dynamically generated euclidean signed distance field for static collision checking. Collision checking for dynamic obstacles is modeled with moving cylinders, increasing the responsiveness to fast-moving agents. Furthermore, we include a Kalman filter motion prediction for moving obstacles into our receding horizon planning, enabling the robot to anticipate possible future collisions. Our experiments\revision{\footnote{Supplementary video: \url{https://youtu.be/_wkqCVz3gdg}}} demonstrate collision-free motions on a quadrupedal robot in challenging indoor environments. The robot handles complex scenes like overhanging obstacles and dynamic agents by exploring motions at the robot's dynamic and kinematic limits.
\end{abstract}

\section{INTRODUCTION}
Legged robots are fast becoming more common in industrial environments~\cite{bellicoso2018jfr}, and it is only a matter of time until we see more of these robots in urban environments~\cite{bbcNewsPandemic}. A necessity for applications in these environments is compliance with \emph{static} environments and \emph{dynamic} agents such as pedestrians moving side by side with the robot. Also, in remote operations where latency to the operator impedes fully supervised teleoperation, an automatic motion planner improves safety and autonomy.

\begin{figure}[t]
    \centering
    \includegraphics[width=0.5\textwidth]{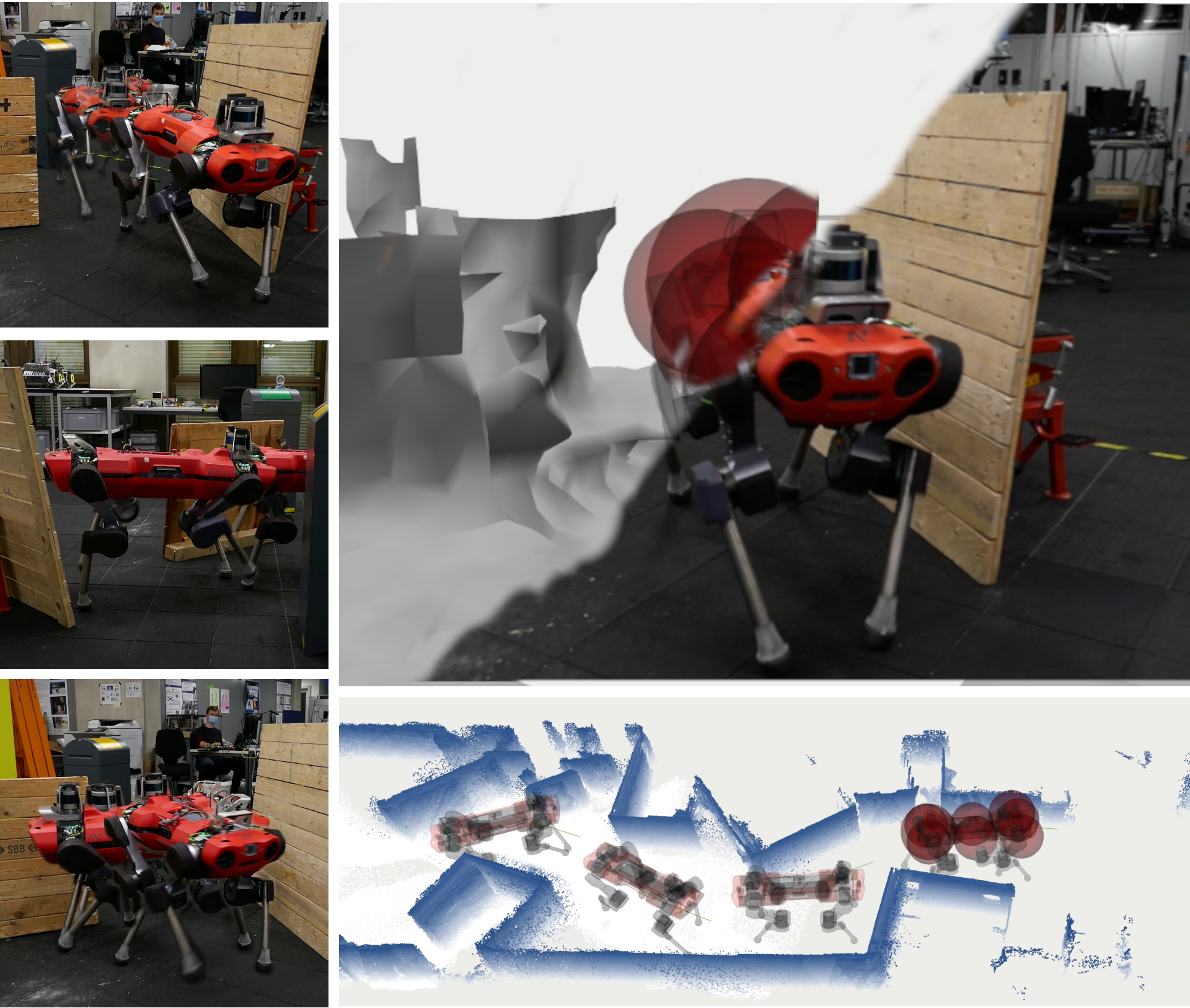}
    \caption{Our quadrupedal robot, \emph{ANYmal}, is navigating collision-free through narrow passages in an office environment. First column: Collision-free locomotion while traversing lateral and rotational through a crowded environment. Top right: \ac{SDF} and sphere decomposition of the robot used to capture the interactions with the environment. Bottom right: Observed trajectory in a narrow corridor while giving a coarse direction input.}
    \label{fig:Teaser}
\end{figure}

The potential advantages of legged robots (as shown in Fig.~\ref{fig:Teaser}) in mobility and versatility compared to wheeled robots come at the cost of higher complexity for the locomotion task. We propose a whole-body \ac{MPC} that simultaneously solves torso and feet motions while being responsive to the environment. Thus, kinematic reachability, dynamic stability, and obstacle avoidance are solved as a whole, leading to an optimal behavior. Moreover, the formulation automatically discovers complex and dynamic motions cumbersome to handcraft through heuristics. The robot successfully anticipates possible future collisions and generates avoidance maneuvers automatically. We show that casting collision avoidance into an optimization problem increases the computational cost of the \ac{MPC} negligibly. But more importantly, it avoids laboriously handcrafting new behaviors.

\subsection{Related Work}
In the following, we categorize existing approaches to collision-free motion planning by their map representation and collision avoidance algorithm and bring them into the context of legged locomotion planning.
\subsubsection{Environment representation}
The environment's representation is a crucial element, and multiple volumetric map representations are available, e.g., octree~\cite{octomap}, height-map~\cite{elevationMapping}, and \ac{ESDF}. \ac{ESDF}s seem to become increasingly crucial in legged robotics as they can deal with the sensor's noise in an inherent fashion~\cite{fankhauser2018robust,buchanan2020perceptive}. Directly storing the distance measurements in the representation enables averaging multiple measurements, thereby filtering the sensor's noise. \emph{Voxblox}~\cite{voxblox} and \emph{FIESTA}~\cite{fiesta} are open-source programs that incrementally update an \ac{ESDF}.

Besides volumetric approaches that are typically aiming for a globally consistent map, local representations are meant to query only local geometric information. One of such representations is \emph{NanoMap}~\cite{nanomap}. Here distances are directly queried from laser scans. It introduces less computational overhead as no explicit map representation is build. On the downside, querying takes longer than on a distance field.  Another approach is to model individual objects of the scene explicitly. For example, the work in~\cite{obstacle_detector} extracts line segments and circles from 2D laser scans and tracks their respective positions and velocities. Typically, objects are approximated by geometric primitives such as planes, spheres, ellipsoids, or polytopes. 
\subsubsection{Optimization-based collision avoidance}
A trajectory optimization method like \emph{\ac{CHOMP}}~\cite{Ratliff2009CHOMP} is widely used for robot motion planning in the presence of obstacles and other constraints. The trajectory costs are invariant to the time parameterization of the trajectory and use pre-computed \ac{SDF}s. In contrast, the work in~\cite{trajopt,Schulman2014Motion} introduces a collision-free trajectory planner, \emph{TrajOpt}. It proposes a sequential convex optimization-based motion planner. Moreover, the algorithm penalizes collisions with a hinge loss. Instead of \ac{SDF}, a convex-convex collision detection is employed to compute minimal translation to avoid collision and directly considers continuous-time safety. The paper in~\cite{Zhang2020Optimization} presents a novel method for reformulating non-differentiable collision avoidance constraints into smooth nonlinear constraints.

The no-collisions constraint can be implemented by constraining the robot's position to be in a safe region. Another approach is to augment the objective with a repulsive potential, typically proportional to the inverse square distance. The former is shown by~\cite{efficientConstraintFormulation} efficiently as a pairwise distance constraint, and the work in~\cite{potentialFieldRelativeVelocity} exemplifies the latter together with a constraint on the relative velocity.
 
The implementation in~\cite{mpcContouringControl} computes the distance between the robot and the environment by discriminating between dynamic obstacles represented by orientable ellipsoids and static obstacles represented by a 2D occupancy grid map. Similarly, the authors in~\cite{sharedSlackVariable} use orientable ellipsoids for moving obstacles and surface normals for static obstacles. A soft constraint with a shared slack variable among all obstacles is introduced; sharing the slack reduces the computational cost when dealing with multiple obstacles.

The recent work in~\cite{johannesVoxblox} shows collision-free planning of end-effector trajectories of a mobile manipulator. The authors add soft inequality constraints with relaxed barrier functions to the \ac{MPC} cost, and a set of collision spheres to approximate the robot.
\subsubsection{Collision-free locomotion planning}
The task of finding the continuous-time motion of legged robots, i.e., the trajectories of the torso and the feet, can be achieved by decomposing the problem or by treating the whole problem as a single task. Decomposed approaches separate the feet and torso planning into two different stages, a common practice to make the collision-free locomotion planning more tractable in terms of the optimization's dimensionality. The full set of possible solutions, however, can not be discovered by such a method, since the individual components do not take into account the full set of physical constraints.

In~\cite{fankhauser2018robust}, the \ac{CHOMP} algorithm generates the shortest collision-free trajectory of the foot's swing trajectory, and the work in~\cite{buchanan2020perceptive} extends the work to collision-free planning of the torso trajectory. The authors generate a \ac{SDF} through a heightmap, and overhanging obstacles are handled through a second 2.5D map of the ceiling. In both approaches, the authors decompose the problem into separate torso and feet trajectory planning. The coordination of each task's solution is one of the main challenges, and heuristics are needed that are cumbersome to handcraft.

By decomposing the locomotion and navigation planner, the planning problem results in similar challenges since a path follower needs to be designed that guides the locomotion planner onto the collision-free path. Recent advancements in robotics competitions are relying on such a decomposition~\cite{hutter2018towards,bouman2020autonomous}. However, these approaches cannot explore highly dynamic motions at their limits, such as using the system's friction and kinematic limits to escape moving obstacles.
\subsection{Contributions}
\revision{We present a MPC for legged robots with a novel way to incorporate static and dynamic environments. We treat }\deleted{
We present a novel \emph{collision-free \ac{MPC}} for legged robots in static and dynamic environments that treats }the whole locomotion planning problem as a holistic optimization\deleted{ problem}, thus simultaneously taking into account the scene, dynamic feasibility, and kinematic reachability. To avoid obstacles, a soft inequality constraint is added as a cost to the switched system. The environment representation is divided into a \emph{static} and \emph{dynamic} scene to increase the responsiveness to fast-moving obstacles. The former uses a dynamically generated \ac{ESDF} for static collision checking, and the latter is based on a moving cylinder representation per obstacle. Our framework extends the capabilities of legged robots in the following ways:
\begin{itemize}
    \item The holistic approach automatically discovers complex, collision-free, and dynamic motions without needless heuristics with minimum computational overhead.
    \item Our whole-body approach enables the robot to explore the full six \ac{DOF} of the torso to handle overhanging obstacles.
    \item Our formulation allows to include the prediction of moving obstacles into our receding horizon planning and enables the robot to anticipate possible future collisions before they take place.
    \item The \ac{MPC} improves the safety and autonomy in narrow passages.
\end{itemize} 

\section{PROBLEM FORMULATION}
At each iteration of the \ac{MPC}, we solve the following optimization problem based on the latest state measurement $\vx_s$ at time $t_s$. We then apply the optimized control policy until a new \ac{MPC} update arrives.
\begin{subequations}
\begin{align}
& \underset{\vu(\cdot)}{\text{minimize}}
&& \int_{t_s}^{t_f} l(\vx(t),\vu(t),t) \,\text{d}t, &&
\label{eq:general_cost} 
\\
& \text{subject to} && \vx(t_s) = \vx_s, &&
\label{eq:initial} 
\\
& && \dot{\vx} = \vf(\vx,\vu,t), &&
\label{eq:general_dynamics} 
\\
& && \vg(\vx,\vu,t) = \mathbf{0} &&
\label{eq:general_equality_constraints}
\\
& && \vh(\vx,\vu,t) \geq \mathbf{0}, &&
\label{eq:general_inequality_constraints} 
\end{align}
\label{eq:op}%
\end{subequations}
where $\vx(t)$ is the state and $\vu(t)$ is the input at time $t$. $l$ is a time-varying running cost. 
The goal is to find a continuous control policy that minimizes this cost subject to the initial condition \eqref{eq:initial}, system dynamics \eqref{eq:general_dynamics}, and general equality \eqref{eq:general_equality_constraints} and inequality \eqref{eq:general_inequality_constraints} constraints. 

In this paper, we employ the \ac{SLQ} algorithm, which is a \ac{DDP} based algorithm for continuous-time systems \cite{Farshidian2017efficient}. \revision{As opposed to Nonlinear Programming (NLP) this approach does not require transcription of the continuous state and input but chooses an adaptive step-size given the desired accuracy.} For \ac{MPC} formulation, we rely on a \deleted{real-time iteration scheme }\revision{real-time iteration scheme where for each MPC cycle, one iteration of SLQ optimization is performed \cite{Diehl2005}.}  In particular we use the \ac{MPC}-\ac{SLQ} approach \cite{Farshidian2017realtime} with feedback policy \cite{grandia2019feedback}. 
\subsection{Blind MPC}
\revision{In this section, we briefly describe the previous \ac{MPC} formulation \cite{Farshidian2017realtime,grandia2019feedback} on which we build our addition for collision avoidance described in Sec. \ref{section:COLLISION-FREE MPC}.}\deleted{In this section, we briefly describe the \ac{MPC} formulation that has been used in blind planning and later has been extended to the collision-free \ac{MPC}. }
\subsubsection{System dynamics} 
We apply our approach to the kino-dynamic model of a quadrupedal robot, which describes the dynamics of a single free-floating body along with the full kinematics for each leg. The state $\vx \in \mathbb{R}^{24}$ and input $\vu \in \mathbb{R}^{24}$ are defined as
\begin{align}
\vx &= ( \vth, \vp, \vom, \vv, \vq ),
\notag
\\
\vu &= ( \vlambda_{ee}^1, \vlambda_{ee}^2, \vlambda_{ee}^3, \vlambda_{ee}^4, \vv_{joint}),
\end{align}
and the equations of motion are given by 
\begin{align}
&\left\{ 
\begin{array}{ll}
		\dot{\vth} = \vT(\vth) \vom \\
		\dot{\vp}  = _W\!\vR_B(\vth) \, \vv \\
		\dot{\vom} = \vI^{-1} \left(  -\vom \times \vI \vom + \sum_{i=1}^4{ {\vr_{ee}^i} \times {\vlambda_{ee}^i}} \right)\\
		\dot{\vv}  = \vg(\vth) + \frac{1}{m} \sum_{i=1}^4{\vlambda_{ee}^i} \\
		\dot{\vq} = \vv_{joint},
\end{array}	 
\right. 
\end{align}
where $_W\!\vR_B$ and $\vT$ are the rotation matrix from the base to the world frame, and the transformation matrix from base local angular velocities to the Euler angles derivatives in the world frame.
$\vg$ is the gravitational acceleration in body frame, $\vI$, and $m$ are the moment of inertia about the \ac{CoM} and the total mass, respectively. The inertia is assumed to be constant and taken at the default configuration of the robot. $\vr_{ee}^i$ is the position of the foot $i$ \ac{w.r.t.} \ac{CoM}. 
$\vth$ is the orientation of the base in \revision{Euler angles\footnote{The reference frame is chosen, such that a gimbal lock appears in the unlikely "backflip" position. }}, $\vp$ is the position of the \ac{CoM} in world frame, $\vom$ is the angular rate, and $\vv$ is the linear velocity of the \ac{CoM}. $\vq \in \mathbb{R}^{12}$ is the joint positions. The inputs of the model are the joint velocities $\vv_{joint} \in \mathbb{R}^{12}$ and end-effector contact forces $\{\vlambda_{ee}^i\}_{i=1}^4$ in the body frame.

\subsubsection{Cost function}
The intermediate cost function is formulated as a quadratic function
\begin{align}
l(\vx, \vu, t) =& \frac{1}{2} \left\Vert \vx-\vx_d  \right\Vert^2_{\vQ}  + \frac{1}{2} \left\Vert \vu-\vu_0 \right\Vert^2_{\vR} ,
 \label{eq:mpc_cost}
\end{align}
where $\vR$ is positive definite and $\vQ$ are positive semi-definite weighting matrices.
$\vx_d$ is the desired state \deleted{consisting of a commanded base pose and twist by the user and a default configuration for the joints.}\revision{computed from a user-defined velocity command and a default configuration for the joints.}
$\vu_0$ is the equilibrium input vector consisting of the statically stabilizing contact forces and zero joint velocities. \revision{Finally, the planning horizon is chosen one second i.e., $t_f - t_s =1$.}

\subsubsection{System constraints}
The constraints capture the different modes of each leg at any given point in time. We assume that the mode sequence is a predefined function of time, and we denote the set of all closed contacts at a given time by $\mathcal{C}$. The resulting constraints are
\begin{equation}
\left\{ 
\begin{array}{lll}
		\vv_{ee}^i(\vx, \vu) = \mathbf{0},
		\quad  &\vlambda_{ee}^i \in \mathcal{Q}(\widehat{\vn}, \mu_c),
		\quad &\text{if $i \in\mathcal{C}$}, \\
		\vv_{ee}^i(\vx, \vu) \cdot \widehat{\vn}  = c(t), \quad  &\vlambda_{ee}^i = \mathbf{0}, \quad  &\text{if $i \notin \mathcal{C}$},
\end{array}
\right.
\end{equation}
where $\vv_{ee}^i$ is the end-effector velocity in the world frame, these constraints ensure that stance legs remain on the ground and swing legs follow a predefined curve, $c(t)$, in the direction of the local surface normal $\widehat{\vn}\in\mathbb{R}^3$ to avoid foot scuffing. 
Furthermore, we enforce zero contact force for swing legs and a friction cone constraint, $\mathcal{Q}$, for stance legs defined by the surface normal and friction coefficient ${\mu_c = 0.7}$.
\section{COLLISION-FREE MPC}
\label{section:COLLISION-FREE MPC}
In this section, we present the collision-free \ac{MPC} formulation that incorporates an additional constraint to achieve collision avoidance. For collision checking, we use a sphere decomposition of the robot and query the distance to the individual spheres' closest obstacle. In the rest of the section, we first present our collision avoidance formulation. We then discuss two different representations of the environment for static and dynamic scenes.

\begin{figure}[t]
\centering
\includegraphics[width=0.4\textwidth]{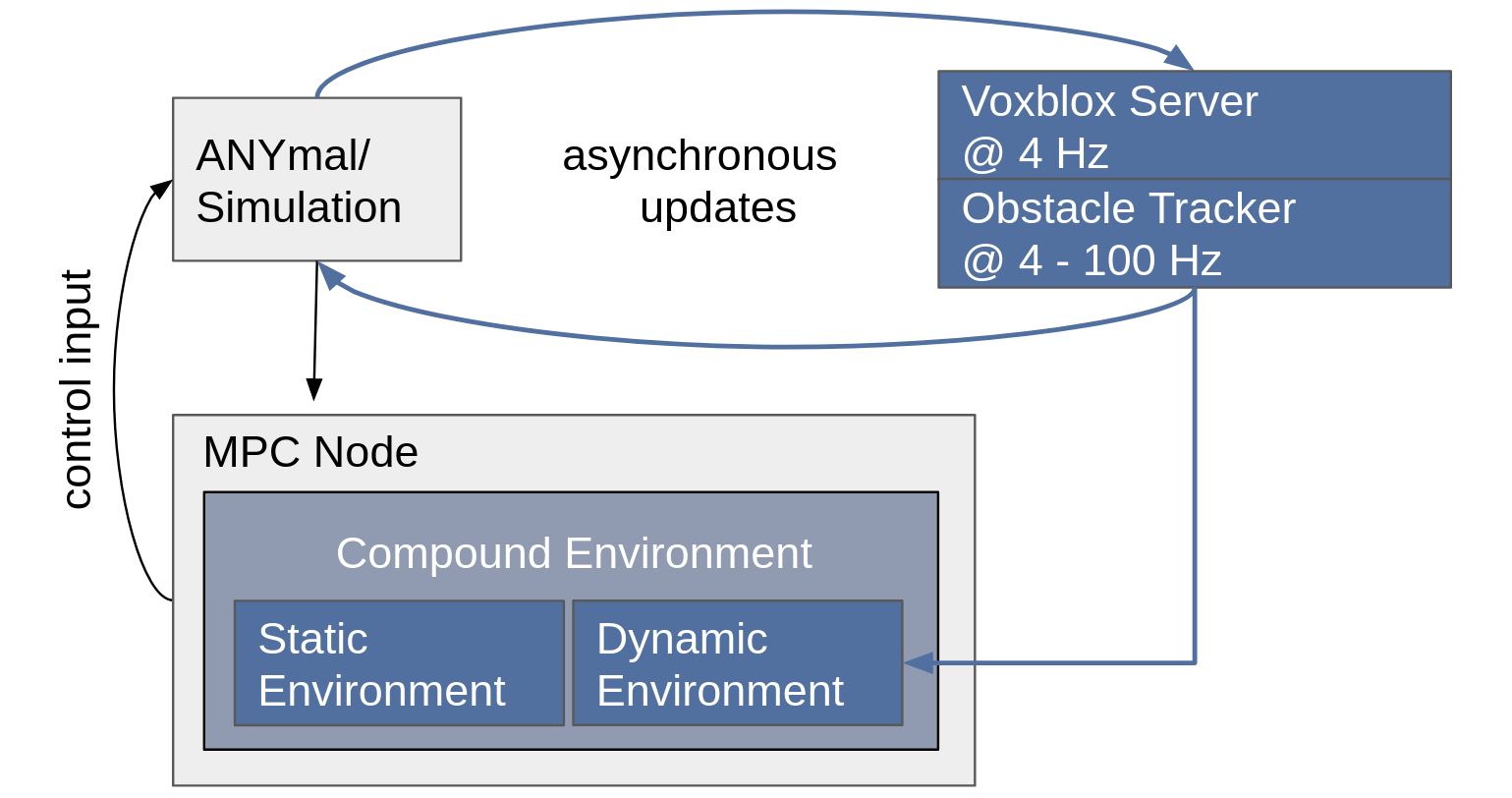}
\caption{Integration of the static and dynamic environment. Two map servers asynchronously update the environment representations from new sensor data. The updates on the representation are sent to the \ac{MPC} node, where they are cached. On each iteration of \ac{MPC}, the most recent cached version is used in the optimization. The augmentation for the collision-free \ac{MPC} is depicted in blue.}
\label{fig:SystemStructure}
\end{figure}
\subsection{Collision Avoidance}
\begin{figure}[t]
\centering
\includegraphics[width=0.5\textwidth]{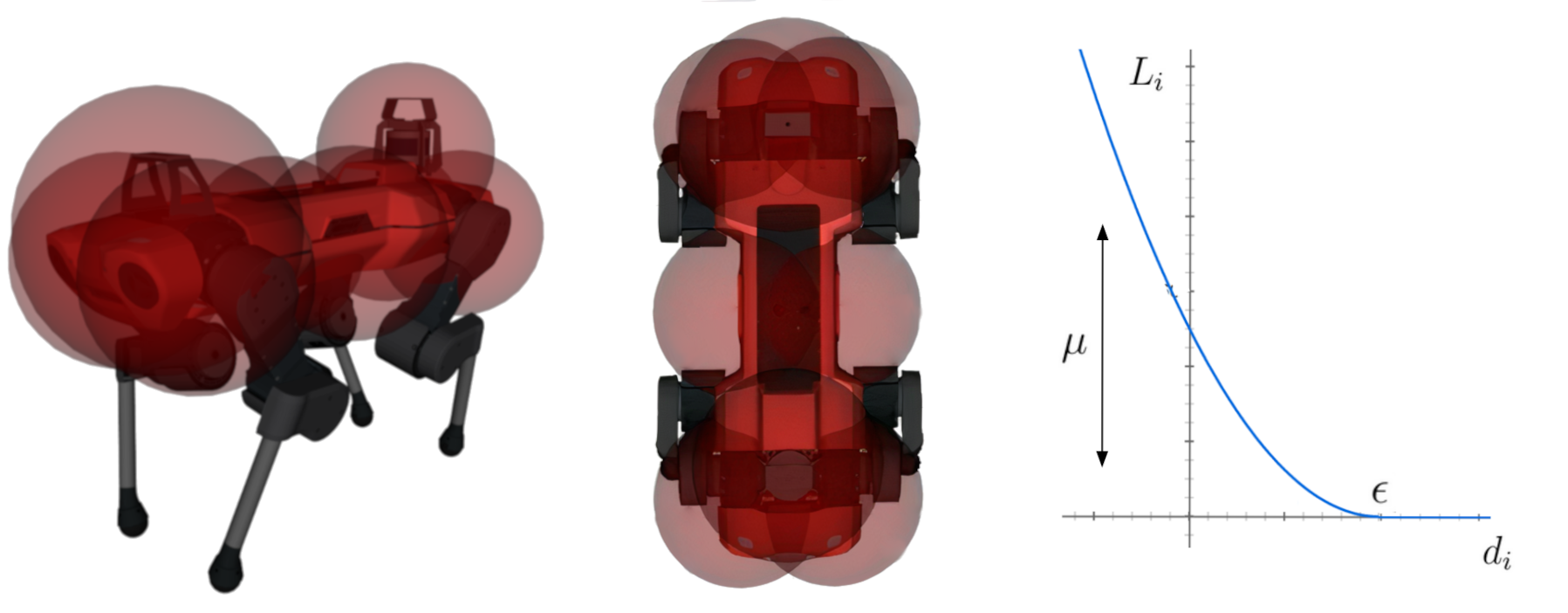}
\caption{Left and center: Sphere decomposition of the robot. Right: A Squared-hinge penalty function is used as a barrier function in the objective. It is parametrized by the scaling parameter $\mu$ and a safety margin $\epsilon$. A distance $d_i < 0$ indicates a collision.}
\label{fig:BarrierFunction}
\end{figure}

Obstacle avoidance is handled as a soft constraint and added to the previously blind \ac{MPC} controller by augmenting the objective with the following penalty function
\begin{align}
l_\text{aug}(\vx, \vu, t) = l(\vx, \vu, t) + l_c(\vx, t),
 \label{eq:mpc_augmentation}
\end{align}
where $l_\text{aug}$ is our augmented intermediate cost function, $l$ the previous intermediate cost from (\ref{eq:mpc_cost}) and $l_c$ our additional penalty function. 
For $l_c$ we use a simple, one-sided quadratic barrier function
\begin{equation}
l_c(\vx, t) = \sum_{i=1}^{p} \mu \cdot \frac{1}{2} \left( max\{0,\epsilon - d_i(\vx, t)\} \right)^2
\label{eq:cost_augmentation} 
\end{equation}
where $p$ is the number of spheres in the sphere decomposition of the body, and $d_i$ is the distance of the sphere $i$ to the closest obstacle. $\mu$ and $\epsilon$ are the scaling and safety-margin parameters illustrated in Fig.~\ref{fig:BarrierFunction}. We used a safety margin of \unit[10]{cm} and \unit[15]{cm} for experiments in simulation and real-world, respectively.

In this work, we decompose the robot's main body by six spheres of uniform size ordered in a two-by-three, largely overlapping grid (see Fig.~\ref{fig:BarrierFunction}). Moreover, two additional spheres are added as collision bodies to account for the front handle and the back's lidar cage. 

To deal with the occasional conflict between the collision avoidance and the position tracking objectives, we clip the position tracking error $\Vert \vp - \vp_d \Vert$ to a maximum threshold of $\unit[15]{cm}$. Doing so, we prevent a sudden pull towards the commanded trajectory once the robot negotiates the blocking obstacle. 

Finally, as the \ac{SLQ} algorithm requires the Hessian of the penalty \ac{w.r.t.} the robot state to be positive semi-definite, we use a \emph{Gauss-Newton} approximation to ensure this property. 
\begin{subequations}
\begin{align}
\frac{\partial^2l_c}{\partial\vx^2} 
&=  
\sum_{i=1}^p {
\mathcal{H}(\epsilon - d_i)
\left(
\mu \frac{\partial d_i}{\partial \vx} \frac{\partial d_i}{\partial \vx}^T + 
\frac{\partial \revision{l_c}\deleted{d_i}}{\partial d_i}\frac{\partial^2 d_i}{\partial \vx^2
} \right)
},
\\
\frac{\partial^2l_c}{\partial\vx^2}
&\approx 
\sum_{i=1}^p {
\mathcal{H}(\epsilon - d_i) 
\left(
\mu \frac{\partial d_i}{\partial \vx} \frac{\partial d_i}{\partial \vx}^T
\right)},
\label{eq:gauss_newton_approximation} 
\end{align}
\end{subequations}
where $\mathcal{H}: \mathbb{R} \to \{0,1\}$ is the Heaviside step function.
\subsection{Environment Representation}
To have the fidelity of a static environment representation and a notion of speed for dynamic obstacles, we combine two environments in a \ac{CCE}. The \ac{CCE} forwards all environmental information, distance and gradient, from the single environment that returns the smallest distance to the \ac{MPC} formulation.

The environment representations are updated asynchronously in separate nodes and updates are sent to the \ac{MPC} node. Our setup is depicted in Fig. \ref{fig:SystemStructure}.  
\subsubsection{Static environment}
We use \emph{Voxblox}~\cite{voxblox} to generate an \ac{ESDF} on a separate \ac{ROS} node, denoted as \emph{Voxblox Server}. 
The Voxblox Server repeatedly sends updates of the computed \ac{ESDF} to the \ac{MPC} node. \revision{Upon receiving a new map, we pre-compute the gradient for each voxel of the distance-field and cache it alongside its distance to make the queries within the MPC optimization faster.}\deleted{On the \ac{MPC} side, the gradients are pre-computed, and both distances and gradients are cached;} This \revision{caching} functionality is taken from~\cite{johannesVoxblox}, although we clip the norm of \revision{this}\deleted{the} gradient to be within its theoretical limits $[0, 1]$ to compensate for errors introduced by finite-difference approximation and the voxelization of the distance field\revision{\footnote{In a static environment, the distance to any object will not change more than the distance directly traveled towards it. In particular, the norm is exactly one, whenever there is exactly one closest object.}}. As the computation time of the distance field update is cubic in the voxel size, we need to trade-off between update rate and resolution. We use an update rate of \unit[4]{Hz} and a voxel size of \unit[10]{cm}. 

At query time, the distance is computed using the first-order approximation from the closest voxel center. Gradients are approximated by the pre-computed gradient of the closest voxel.
\subsubsection{Dynamic environment}
Although the use of an \ac{ESDF} representation seems best suited for our requirements, it has one limitation that is hard to overcome - it is inherently static. Consequently, we can not incorporate predictions of the state in the future into the receding time horizon of the controller. Instead, we use a simplistic, cylindrical representation for dynamic obstacles. By introducing an object-based representation for moving obstacles, we can track their trajectories and predict their motions. For this, we make use of an open-source \ac{ROS} package \emph{obstacle\_detector}~\cite{obstacle_detector} that takes ordered, two-dimensional laser-scans. \revision{We use their \emph{obstacle\_publisher} auxiliary node to publish a set of virtually detected circles for the experiments in simulation. The state of an obstacle is the two-dimensional position, velocity, and radius. We use a constant-velocity model to predict the future location of these moving obstacles and integrate it into our receding horizon \ac{MPC} formulation. }\deleted{The package then extracts line segments and circles and tracks the latter using a Kalman filter. We take this cylindrical representation of obstacles, who's state is given by their two-dimensional position, velocity, and radius. This allows us to extrapolate their position in the future using a constant velocity model and utilize the prediction in the receding horizon framework. }
\section{EXPERIMENTAL RESULTS AND DISCUSSION}
\begin{figure*}[t]
    \centering
    \includegraphics[width=\textwidth]{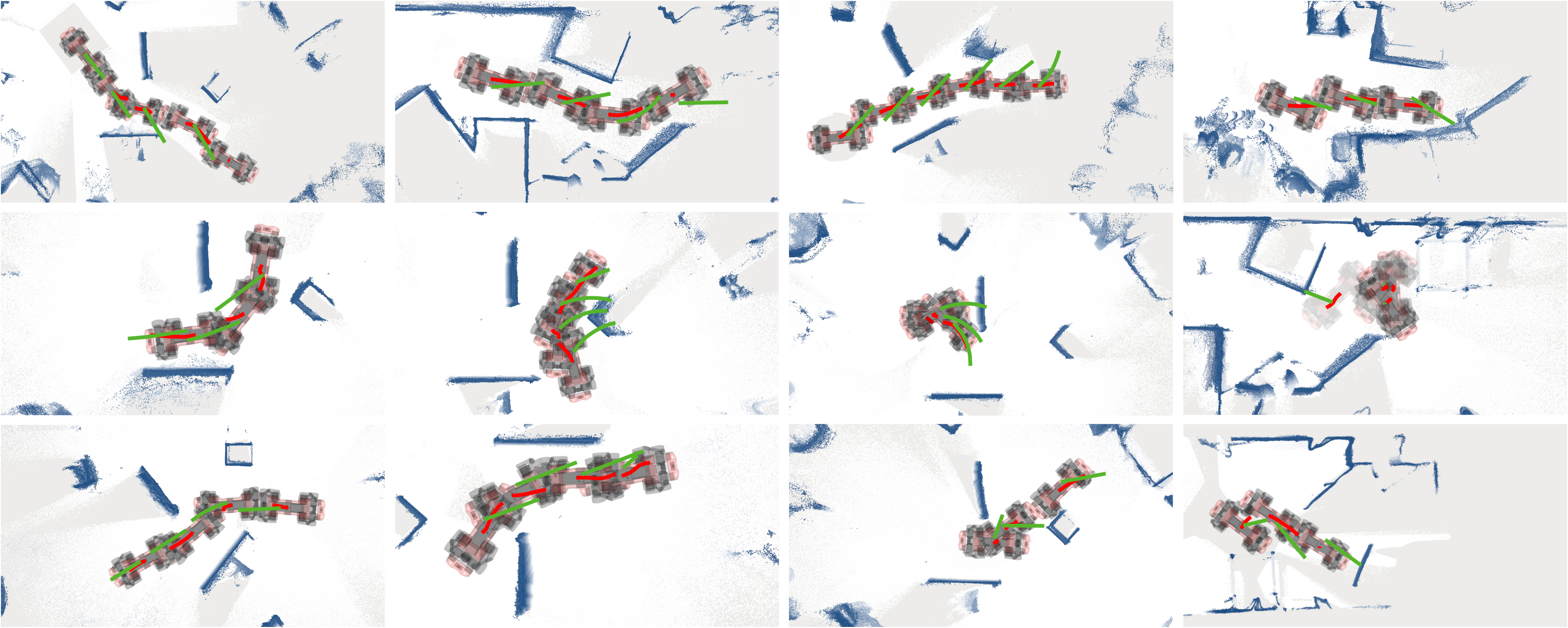}
    \caption{Collection of different emerging behaviors during real-world experiments. At each snapshot the commanded (green) and optimized (red) trajectories are shown. First row, lateral commands: The robot adjusts its yaw and position to avoid obstacles along the commanded trajectory. Middle row, rotational commands: The center of rotation is adjusted dynamically to both follow the desired command closely but not collide with the environment. Bottom row: The robot followed more diverse inputs while avoiding obstacles.}
    \label{fig:ResultsOverview}
\end{figure*}
We show that our formulation discovers complex behaviors leveraging the full dynamic and kinematic capabilities of the robot. Experiments have been carried out both in simulation and real-world on our quadrupedal robot, ANYmal \cite{hutter2017anymal}. 
In this section, we first study the computational overhead incurred by our perceptive formulation compared to the blind version. We then evaluate the performance of our controller in different environments with static and dynamic obstacles.
\subsection{Computational Cost}
\begin{figure}[t]
    \centering
     \includegraphics[width=0.5\textwidth]{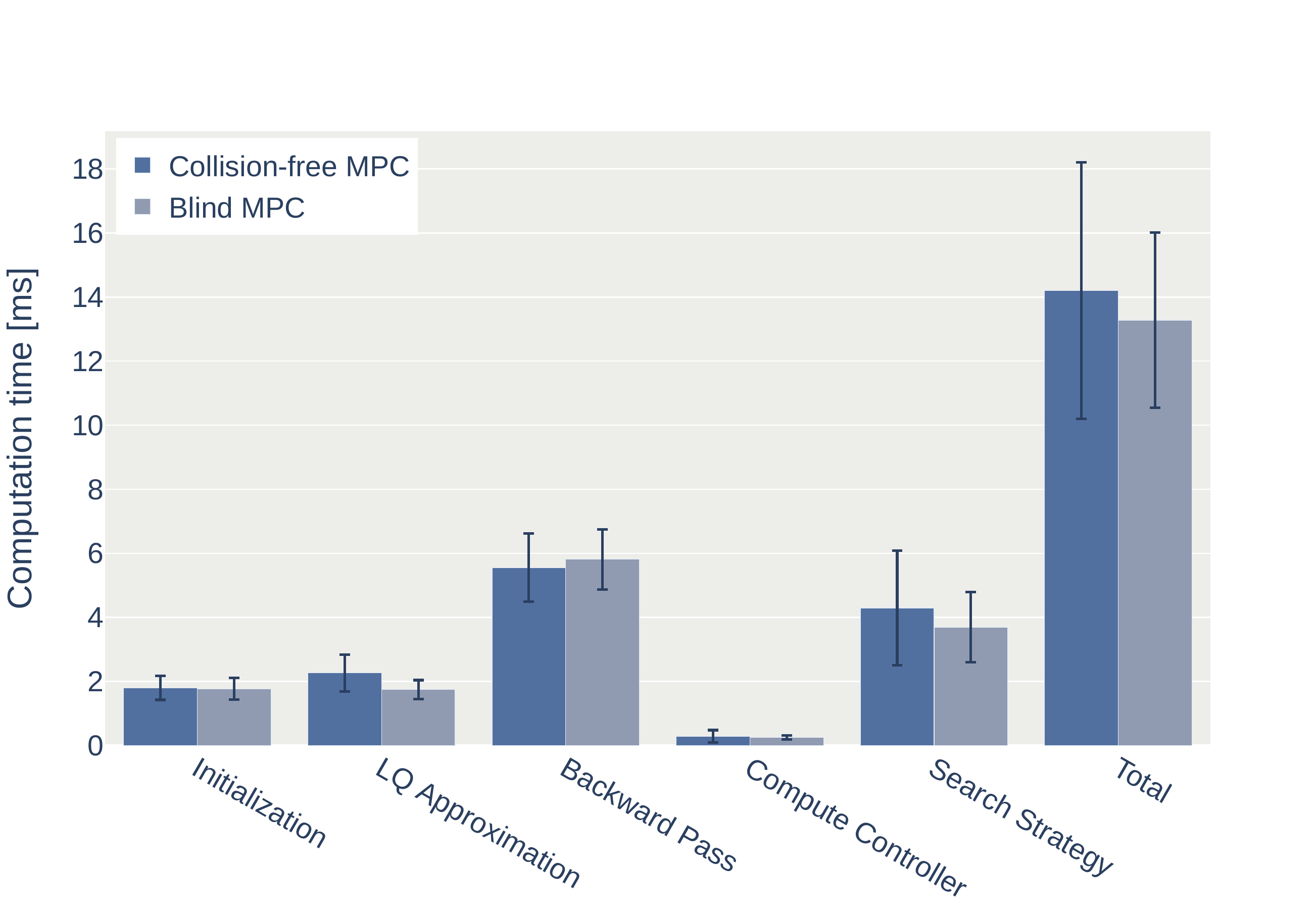}
    \caption{Computational comparison of the collision-free and blind \ac{MPC}. The serial computation time of the solver is split up into its main parts. Bars show the average of over 1000 solver iterations. Black error bars indicate one standard deviation. On average, the solving time increased only marginally by \unit[0.93]{ms}. For this benchmarking, we use AMD Ryzen 5 2600x, \unit[16]{Gb} DDR4 RAM, 64bit.}
    \label{fig:Benchmark}
\end{figure}

Benchmarking our controller against a blind \ac{MPC} reveals that the computation time is only slightly increased. \revision{While operating at a frequency of 20 Hz, our }\deleted{Our }collision-free \ac{MPC} formulation introduces a computational overhead in the backward pass and in searching a strategy amounting to a total increase of \unit[6]{\%} or \unit[0.93]{ms}. At the same time, it breaks down the complexity of the control structure and engineering effort required for connecting inherently-dependent but independently-designed path navigation and tracking modules. More detailed timings are shown in Fig.~\ref{fig:Benchmark}.
\subsection{Whole-Body Planning and Look-Ahead}
\begin{figure}[t]
    \centering
    \includegraphics[width=0.5\textwidth]{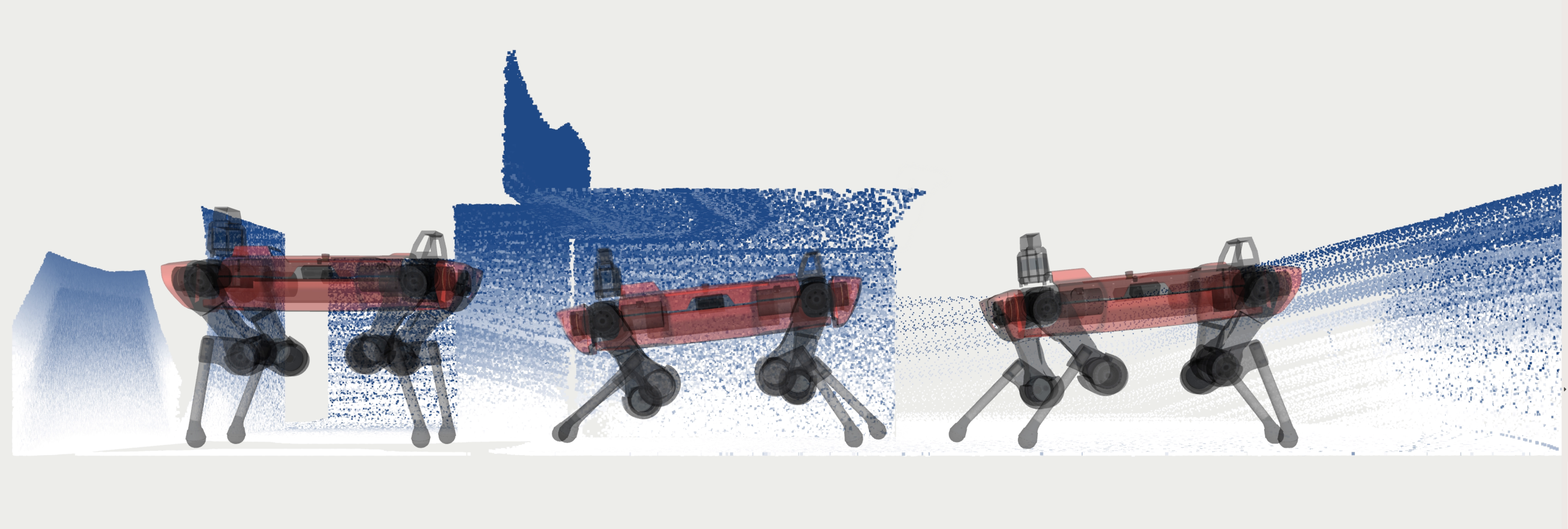}
    \caption{ANYmal ducking under the overhanging obstacle. While commanded to go forward, the robot automatically adjusts its base height and orientation to avoid a collision with the overhanging obstacle. Afterwards, the robot adjusts to its nominal posture.}
    \label{fig:Ducking}
\end{figure}

\begin{figure}[t]
    \centering
    \includegraphics[width=0.5\textwidth]{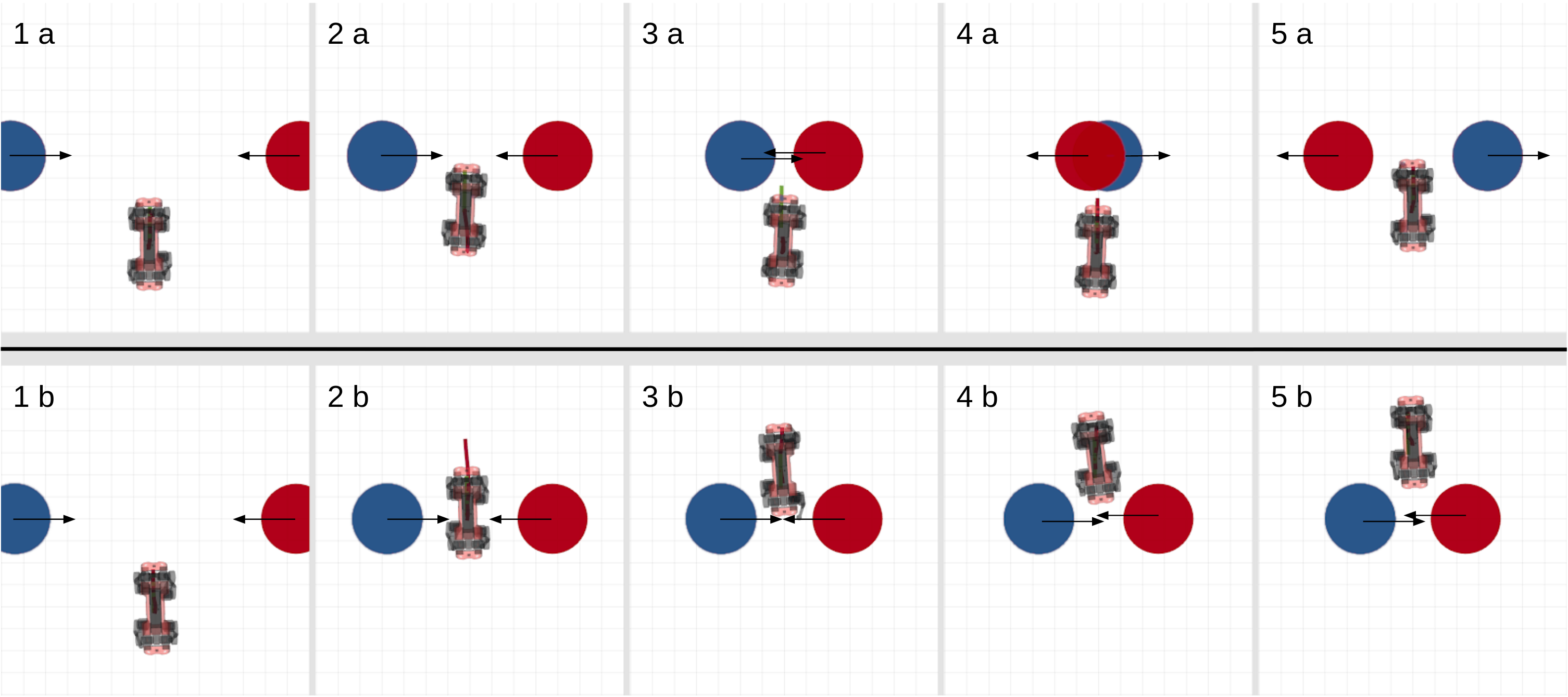}
    \caption{Sequences of dynamic obstacle avoidance maneuvers. Top, left to right: Obstacles approaching at \unit[0.5]{m/s}. The robot is \revision{trotting in place }waiting for the obstacles to pass, although continuously commanded to go ahead. Note that at (2a) - (4a), the robot even moves backward. Bottom, left to right: Obstacles approaching at \unit[0.2]{m/s}. The robot accelerates (2b) to pass through the obstacles to then decelerate (4b), (5b) to be again aligned with the desired trajectory.}
    \label{fig:DynamicObstacles}
\end{figure}

Our formulation shows its potential best while locomoting through narrow passages and more complex maneuvers such as ducking and dynamic collision avoidance.
Thereby, the whole body optimization leverages the inter-dependencies between robot dynamics and collision-free trajectories. We carried out two evaluations in simulation, for ducking under overhanging obstacles (see Fig.~\ref{fig:Ducking}) and avoidance of dynamic obstacles (see Fig.~\ref{fig:DynamicObstacles}) \revision{. For those experiments, the robot was given only a constant forward velocity command and no explicit final position was given or updated to avoid the obstacles. Moreover we carried out several experiments in real world (see Fig.~\ref{fig:ResultsOverview} and the \href{https://youtu.be/_wkqCVz3gdg}{\UrlFont{attached video}}).}\deleted{, and several experiments in real-world (see Fig.~\ref{fig:ResultsOverview}).}
\subsubsection{Overhanging obstacles}
The robot automatically ducks under overhanging obstacles and returns to its nominal posture once the obstacle is passed. In contrast to the 2.5D height map in~\cite{buchanan2020perceptive}, our environment representation based on the \ac{SDF} can handle the full 3D space, including overhanging obstacles.
\subsubsection{Dynamic obstacles}
As mentioned before, our collision-free \ac{MPC} allows the integration of the prediction of moving obstacles into its receding horizon planning. To showcase this property, we devise two tests where ANYmal should pass through two approaching obstacles from opposite sides. In both cases, the robot is commanded forward with constant velocity. In the first case, the objects are moving faster (\unit[0.5]{m/s}) than in the second case (\unit[0.2]{m/s}), see Fig.~\ref{fig:DynamicObstacles}. In both cases, the robot automatically anticipates the timing when it is possible to surpass the moving obstacles without collision. We can demonstrate that the robot prefers to surpass these obstacles when the dynamic obstacles move at a slower speed. At higher speeds, the robot \revision{trots in place till the obstacles have passed }\deleted{waits for the obstacles to pass }before continuing its path. This highlights that in this formulation, depending on the scene and system's limits (in this case, friction cone limits and the actuation cost), different behavior can emerge within the same optimization problem.
\subsection{Hardware Experiments}
\begin{figure}[t]
    \centering
    \includegraphics[width=0.5\textwidth]{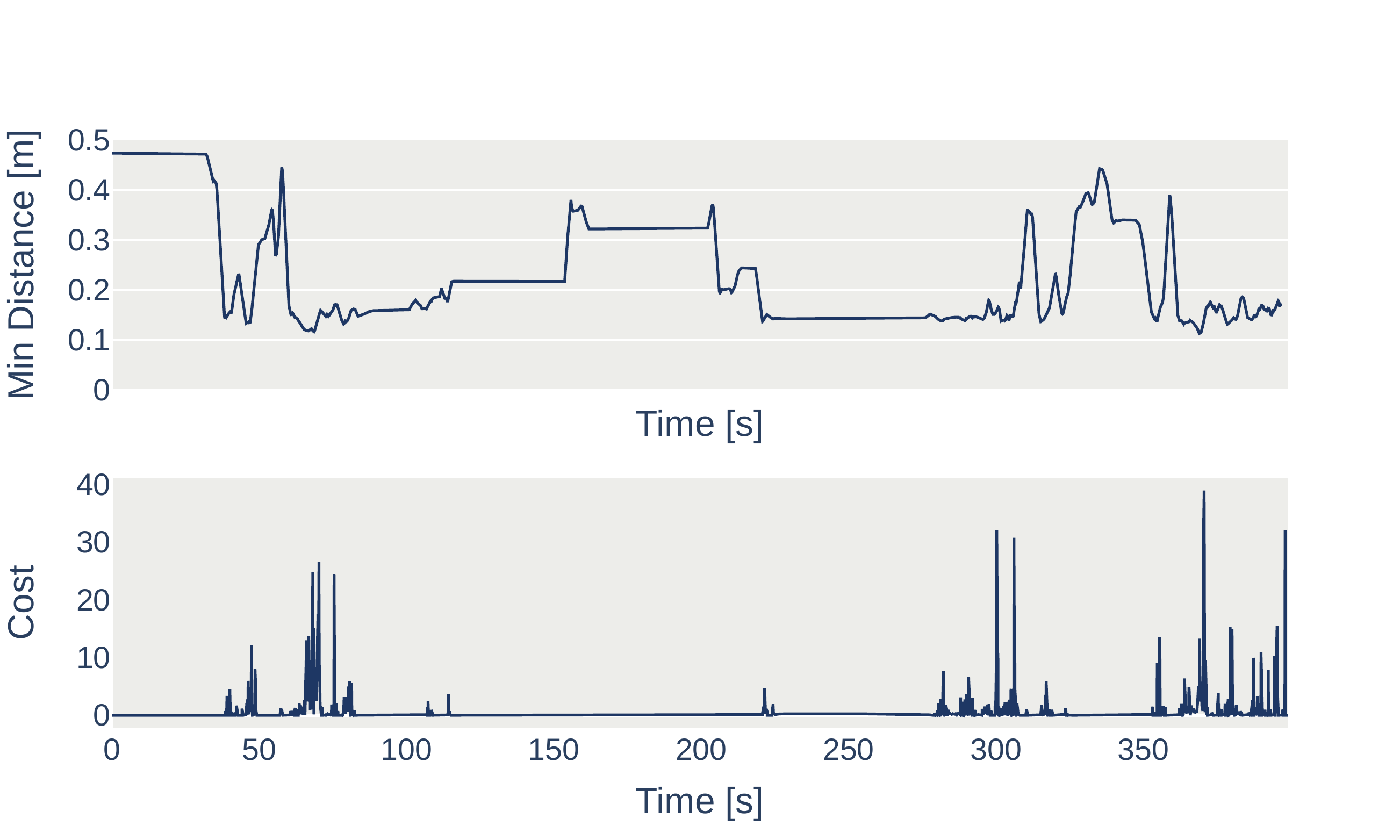}
    \caption{Top: Distance of the robot to the closest obstacle. We observe that the distance stays well above zero. Bottom: Distance penalty associated with the current state. For distances above \unit[0.15]{m} the cost is zero. A Moving average is applied over a time window of \unit[1.6]{s}.}
    \label{fig:CollisionMetrics}
\end{figure}
Several real-world experiments have shown that our planner performs comparably \ac{w.r.t.} the simulation results on real hardware. We observed motions like sidestepping, aligning to walls, and backward movement in order to execute a commanded turn.
Fig.~\ref{fig:ResultsOverview} gives a qualitative overview of the observed motions. While navigating through artificially narrowed down office floors (Fig. \ref{fig:Teaser}) for over five minutes, we had only one minor collision with a very slim obstacle that was not properly mapped. Fig.~\ref{fig:CollisionMetrics} shows the minimum distance and its associated cost for this experiment. 
Encountered map inconsistencies caused by an odometry drift were partially solved by a map update rate of \unit[4]{Hz} and trusting more on the recently received sensor data. The collision-free \ac{MPC} increases the safety factor of the operation and helps the operator go through cluttered space.
\section{CONCLUSIONS}
We introduce a collision-free \ac{MPC} for legged robots that enables our four-legged robot, ANYmal, to navigate cluttered environments and deal with dynamic obstacles. In contrast to existing approaches, the problem is solved as a holistic optimization problem that considers the collision-free space, dynamic feasibility, and kinematic constraints over a receding horizon. With this formulation, we can include the prediction of moving obstacles, which enables the robot to avoid future collisions before they occur. To achieve collision avoidance, we add a relaxed barrier function to the switched system's \ac{MPC} formulation without increasing the problem's computational complexity. ANYmal manages to automatically discover complex motions, avoiding cluttered scenes with overhanging and moving obstacles. Also, the real-world tests verify that the algorithm can cope with noisy data from exteroceptive sensors. All in all, the collision-free \ac{MPC} improves the safety and autonomy of the robot.

In future work, we plan to incorporate collision-free planning of the swing feet trajectory to negotiate rough terrain. Furthermore, we aim to employ more sophisticated prediction strategies \revision{\cite{hoeller2021learning}} for moving obstacles based on a history of object motion making a better prediction of their future path beyond the first-order model used in this work.
\section*{Acknowledgments}
\small{
This research was supported by the Swiss National Science Foundation (SNSF) through project 166232, 188596, the National Centre of Competence in Research Robotics (NCCR Robotics), and the European Union’s Horizon2020 (grant agreement No. 852044). Moreover, this work has been conducted as part of ANYmal Research, a community to advance legged robotics.

We appreciate the insightful discussion and support from Marco Tranzatto, ETH Z\"urich.
}

\bibliographystyle{bibtex/myIEEEtran} 
\bibliography{bibtex/IEEEabrv,bibtex/IEEEexample,bibtex/bibliography}

\end{document}